\documentclass[10pt,twocolumn,letterpaper]{article}

\usepackage[pagenumbers]{cvpr} 

\usepackage{pifont}
\usepackage{colortbl}

\definecolor{cvprblue}{rgb}{0.21,0.49,0.74}
\usepackage[pagebackref,breaklinks,colorlinks,allcolors=cvprblue]{hyperref}
\usepackage{amsmath}

\def\paperID{} 
\def\confName{}
\def\confYear{}

\title{Monocular Models are Strong Learners for Multi-View Human Mesh Recovery}

\author{
Haoyu Xie$^{1}$ \quad
Shengkai Xu$^{1}$ \quad
Cheng Guo$^{1}$ \quad
Muhammad Usama Saleem$^{1}$ \quad
Wenhan Wu$^{1}$ \\
Chen Chen$^{2}$ \quad
Ahmed Helmy$^{1}$ \quad
Pu Wang$^{1}$ \quad
Hongfei Xue$^{1^*}$ \\
$^{1}$University of North Carolina at Charlotte \quad
$^{2}$University of Central Florida \\
\\
{\tt\small \{hxie3, sxu7, cguo3, msaleem2, wwu25, ahmed.helmy, pu.wang, hongfei.xue\}@charlotte.edu} \\
{\tt\small chen.chen@ucf.edu}
\vspace{-20pt}
}

\begin{document}
\maketitle
\begingroup
\renewcommand{\thefootnote}{\fnsymbol{footnote}}
\footnotetext[1]{Corresponding author.}
\endgroup
\begin{abstract}
Multi-view human mesh recovery (HMR) is broadly deployed in diverse domains where high accuracy and strong generalization are essential. 
Existing approaches can be broadly grouped into geometry-based and learning-based methods. 
However, geometry-based methods (e.g., triangulation) rely on cumbersome camera calibration, while learning-based approaches often generalize poorly to unseen camera configurations due to the lack of multi-view training data, limiting their performance in real-world scenarios.
To enable calibration-free reconstruction that generalizes to arbitrary camera setups, we propose a training-free framework that leverages pretrained single-view HMR models as strong priors, eliminating the need for multi-view training data.
Our method first constructs a robust and consistent multi-view initialization from single-view predictions, and then refines it via test-time optimization guided by multi-view consistency and anatomical constraints.
Extensive experiments demonstrate state-of-the-art performance on standard benchmarks, surpassing multi-view models trained with explicit multi-view supervision.
\end{abstract}    

\section{Introduction}
\label{sec:intro}

Multi-view human mesh recovery (HMR), which reconstructs accurate 3D human pose and shape from synchronized camera views, is a fundamental capability for high-fidelity human-centric perception.
Compared with single-view methods~\cite{goel2023humans,patel2025camerahmr,dwivedi2024tokenhmr, li2022cliff,saleem2025genhmr} that inherently suffer from depth ambiguity, occlusions, and viewpoint-dependent uncertainty, multi-view approaches greatly improve reliability and geometric accuracy by leveraging complementary spatial information across cameras.
At the same time, they offer a practical, accurate, and markerless alternative to traditional marker-based motion capture systems~\cite{shuai2021easymocap,Pagnon_2022_JOSS,Pagnon_2022_Accuracy,Pagnon_2021_Robustness}.
These advantages make multi-view human mesh recovery essential for downstream applications such as fine-grained behavior understanding~\cite{ragusa2024enigma,singh2016towards}, human digitalization~\cite{hassani2021human, chan2022emergence,toivonen2019human}, and immersive AR/VR experiences~\cite{moloney2018affordance,neo2021designing,zhang2023survey}.

Early multi-view HMR methods~\cite{huang2017towards, liang2019shape, shin2020multi, chun2023learnable} have demonstrated the benefit of aggregating information from multiple viewpoints to improve reconstruction accuracy.
Volumetric approaches~\cite{shin2020multi, chun2023learnable} typically fuse multi-view 2D features into a unified 3D feature volume via triangulation, explicitly leveraging geometric constraints during fusion.
This design enables strong generalization across different camera configurations. However, this camera calibration-requiring design, as well as the inherent quantization errors introduced by discrete volumetric representations, limits their scalability in practical settings.
To further improve reconstruction accuracy, recent methods \cite{jiang2022multi,li2024human} adopt end-to-end learning frameworks that fuse multi-view features, achieving strong performance in modeling human pose and shape.
However, existing multi-view datasets provide only a narrow range of camera configurations~\cite{h36m_pami, mono-3dhp2017}.
As a result, the learned fusion modules inevitably memorize dataset-specific geometric priors tied to those limited camera settings.

This baked-in implicit bias severely restricts generalization: although these models perform well under training conditions, their accuracy drops sharply when evaluated on unseen camera settings.
A representative example is U-HMR~\cite{li2024human}, whose feed-forward architecture aggregates image features to directly regress the 3D mesh. The model tends to memorize the 2D-to-3D correspondences observed during training, leading to substantial degradation under novel camera configurations, as demonstrated in~\cite{matsubara2025heatformer}.
A similar issue arises in MVP~\cite{NEURIPS2021_6da9003b}, where \cite{liao2024multiple} shows that its performance collapses to near-zero when tested on new camera settings, indicating that it encodes specific camera parameters rather than learning robust multi-view reasoning.
%

In parallel, single-view HMR methods~\cite{goel2023humans,patel2025camerahmr,dwivedi2024tokenhmr, li2022cliff,saleem2025genhmr} have witnessed remarkable progress. A pivotal advantage of these single-view methods is their access to massive-scale training datasets~\cite{h36m_pami,mono-3dhp2017,lin2014microsoft, andriluka14cvpr,tripathi2023ipman,murray2012ava,wu2017ai}, including extensive synthetic datasets~\cite{mahmood2019amass,Black_CVPR_2023,vonMarcard2018}. This data advantage enables them to learn a powerful and reliable prior of the human body with diverse camera viewpoints and settings, avoiding the camera-specific generalization issues that plague multi-view learning.
Landmark works have steadily advanced the state of the art: HMR2.0~\cite{goel2023humans} established a pure transformer-based baseline, TokenHMR~\cite{dwivedi2024tokenhmr} replaced the regression head with a well-trained codebook for accurate and more plausible outputs, CLIFF~\cite{li2022cliff} incorporated bounding box information to improve camera-space estimation, and CameraHMR~\cite{patel2025camerahmr} integrated perspective camera model to mitigate projection error. However, these single-view solutions are inherently constrained by the fundamental ambiguities of monocular reconstruction, namely depth uncertainty and occlusion. 

At this point, pure learning-based multi-view fusion modules lack generalization due to the scarcity of camera settings in the training data, but single-view methods provide a powerful and generalizable human mesh prior learned from large-scale data.
This leads us to ask the following key question: \textit{Is it possible to construct a multi-view HMR system that relies solely on single-view training, yet delivers high accuracy and robustness across arbitrary camera setups?}

To address this question, we introduce a framework that injects single-view priors into multi-view reconstruction via iterative optimization. Our approach is built on two key insights: (1) modern single-view HMR models provide sufficiently strong and generalizable priors to serve as reliable foundations for 3D reconstruction, and (2) robust generalization to arbitrary camera configurations emerges only when multi-view fusion explicitly reasons about 3D geometry, rather than relying on purely data-driven models that implicitly memorize camera settings.
To operationalize these principles, our framework directly initializes an optimizable ``virtual view'' representation from per-view estimates using pre-trained single-view models without any multi-view training, then optimizes this representation and per-view estimates via Test-time adaptation (TTA) guided by 2D anatomical landmarks, multi-view consistency, and single-view prior regularization.
To conclude, our contributions are summarized as follows:

\begin{itemize}
\item We propose a multi-view TTA framework that iteratively refines single-view priors guided by multi-view information and 2D anatomical landmarks, enabling accurate and robust mesh recovery without requiring any multi-view training data.
\item Through comprehensive ablation studies, we systematically explore key design choices of our TTA paradigm, including the selection of optimizable components, the formulation of guidance, and the hyper-parameter weights.
\item Our method achieves state-of-the-art performance on benchmark datasets even compared to fully supervised multi-view approaches, demonstrating superior generalization to novel camera configurations.
\end{itemize}

\begin{figure*}[!t]
    \centering
    \includegraphics[width=0.9\linewidth]{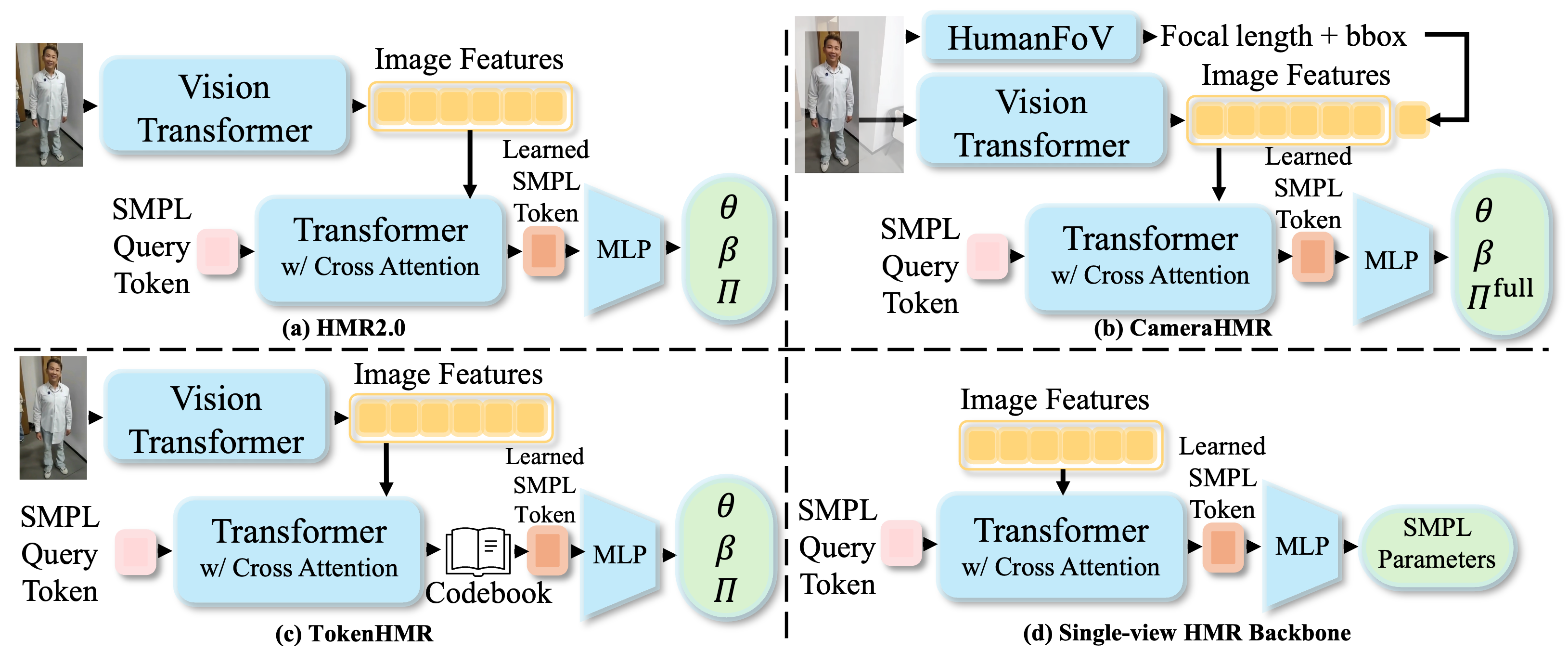}
    \caption{The model architectures of (a) HMR2.0, (b) CameraHMR, (c) TokenHMR, and (d) a summary of these generic single-view transformer-based HMR models.
    The optimizable components can be selected from three representative groups: tokens (pink/orange), model parameters (blue), or explicit SMPL parameters (green).}
    \label{fig:components}
\end{figure*}
\section{Related Work}
\subsection{Single-view Human Mesh Recovery}
\vspace{-3pt}
Single-view human mesh recovery~\cite{bogo2016keep,kolotouros2019learning,moon2022neuralannot,zhang2021pymaf,sengupta2023humaniflow} has advanced rapidly due to large-scale datasets and strong parametric priors.
Early work such as HMR~\cite{kanazawa2018end} introduced the standard top-down pipeline that regresses SMPL parameters from image features via iterative refinement. CLIFF~\cite{li2022cliff} improved global pose estimation by restoring lost spatial cues through bounding-box conditioning and full-image reprojection. HMR2.0~\cite{goel2023humans} replaced CNN encoders with a ViT-H/16 backbone, demonstrating the benefit of high-capacity transformers for mesh prediction. TokenHMR~\cite{dwivedi2024tokenhmr} addressed camera-model mismatch by proposing adaptive loss scaling and a discrete pose-token representation learned from motion capture data. Recent extensions such as CameraHMR~\cite{patel2025camerahmr} further refine camera modeling to reduce projection errors. Together, these methods provide powerful and highly generalizable single-view priors, but remain fundamentally limited by monocular ambiguitie. 

\vspace{-5pt}
\subsection{Multi-view Human Mesh Recovery}
\vspace{-3pt}
Multi-view HMR leverages complementary observations across cameras to resolve depth ambiguity and occlusion that fundamentally limit single-view approaches. Early geometric methods rely on triangulation or volumetric fusion~\cite{iskakov2019learnable,shin2020multi}, lifting multi-view 2D cues into a discretized 3D space. Although explicitly encoding geometry and generalizing across camera setups, these approaches suffer from voxel quantization and are brittle under occlusion or imperfect calibration.
Subsequent learning-based pipelines aim to fuse view-specific features directly. View-by-view refinement~\cite{liang2019shape} propagates predictions across views but underuses camera parameters and often accumulates cross-view inconsistencies. Feed-forward transformer architectures~\cite{jiang2022multi,li2024human} aggregate multi-view image features to regress SMPL parameters; however, their fusion remains implicit and tends to absorb dataset-specific camera priors, leading to poor generalization to unseen configurations.
Recent iterative refinement methods~\cite{jia2023delving,matsubara2025heatformer} revisit optimization-style updates for multi-view HMR. Pixel-aligned fusion~\cite{jia2023delving} reprojects intermediate meshes to extract feedback signals, while HeatFormer~\cite{matsubara2025heatformer} formulates multiview alignment as heatmap matching via a transformer-based neural optimizer. These approaches improve robustness to occlusion and camera variation, however, their methods performance degrades in calibration-free mode. 
HeatFormer relies on epipolar geometry in AdaFuse~\cite{zhang2021adafuse} module to align cross-view heatmaps; without reliable calibration, cross-view information cannot be effectively aggregated. Similarly, PaFF~\cite{jia2023delving} depends on camera parameters for pixel-aligned feedback, orientation alignment, and scale recovery. 
Moreover, both methods require multi-view training data and may implicitly encode dataset-specific camera layouts, limiting their generalization to unseen configurations.

\section{Method}

\subsection{Overview}

Our goal is to recover a robust and consistent 3D human mesh from multiple camera views while generalizing to unseen camera configurations, without relying on multi-view training data.
To achieve it, we build on pretrained single-view HMR models and refine their predictions at inference time by optimizing a small set of variables under multi-view constraints and 2D anatomical clues.
However, this design brings to two key questions: \emph{(1) what to optimize}—which parts of a pretrained single-view model should be updated at test time to enable effective adaptation; and \emph{(2) how to optimize}—which constraints should guide the updates to enforce cross-view consistency and anatomical plausibility.
We answer these questions by (i) analyzing different choices of optimizable components and their initialization in the multi-view setting, and (ii) introducing a set of test-time objectives that drive refinement toward coherent solutions.

The remainder of this section is organized as follows. Sec.~3.2 briefly reviews the required preliminaries. Sec.~3.3 outlines the overall inference-time optimization pipeline. Sec.~3.4 studies the choice and initialization of optimizable components. Sec.~3.5 presents the test-time objectives used to guide refinement.



\subsection{Preliminaries}

\begin{figure*}[!t]
    \centering
    \includegraphics[width=1\linewidth]{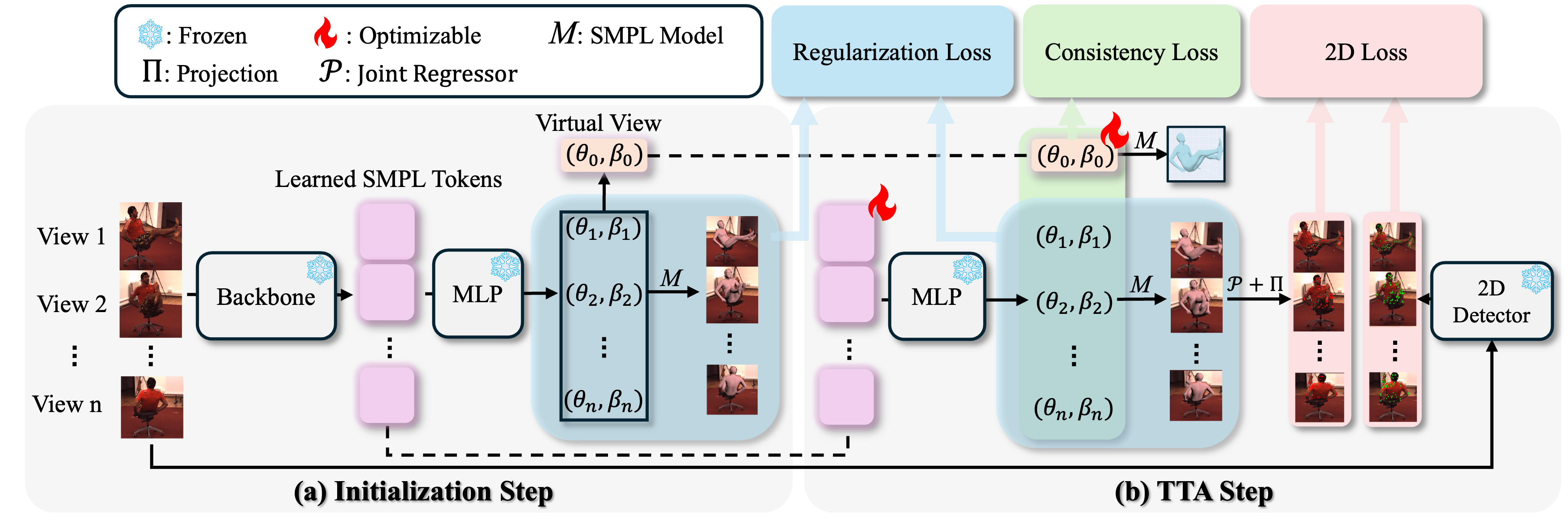}
    \caption{Test-time adaptation (TTA) framework for multi-view human mesh recovery. The pipeline is composed of two stages: 
    (a) aggregating single-view SMPL estimates into a set of SMPL parameters initialization termed \emph{virtual view}; (b) Test-time adaptation refines single-view optimizable token and virtual view SMPL parameters using 2D Loss, multi-view consistency Loss, and regularization loss.}
    \label{fig:monomv_architecture}
\end{figure*}

\noindent \textbf{SMPL Model.}
The SMPL model~\cite{loper2015smpl} represents the human body using a low-dimensional parametric formulation. It is controlled by pose parameters $\theta \in \mathbb{R}^{72}$, encoding the relative 3D rotations of 24 joints, and shape parameters $\beta \in \mathbb{R}^{10}$, capturing person-specific body shape variations.
Given $(\theta, \beta)$, SMPL outputs a posed human mesh $V = M(\theta, \beta) \in \mathbb{R}^{6890 \times 3}$ using linear blend skinning. The corresponding 3D joint coordinates $J_{3D} \in \mathbb{R}^{K \times 3}$ are obtained by applying a fixed linear joint regressor to the mesh. \textbf{In our framework}, SMPL model is used for generating human mesh from SMPL parameters.

\noindent \textbf{Single-view HMR Backbones.}
Recent transformer-based single-view HMR models directly regress the SMPL parameters $(\theta, \beta)$ and camera coefficients $\Pi$ from a single RGB image.
HMR2.0~\cite{goel2023humans} (\cref{fig:components}(a)) establishes a strong baseline by adopting a Vision Transformer (ViT~\cite{dosovitskiy2020image}) encoder to extract image features and a transformer decoder that operates on learnable SMPL tokens, which are finally mapped to $(\theta, \beta)$ via an SMPL regression head.
Building upon this architecture, CameraHMR~\cite{patel2025camerahmr} (\cref{fig:components}(b)) improves reconstruction by explicitly modeling the perspective camera. It predicts the camera field-of-view and introduces camera-aware tokens into the transformer to enable more accurate projection.
TokenHMR~\cite{dwivedi2024tokenhmr} (\cref{fig:components}(c)) explores a complementary direction by replacing continuous pose regression with a tokenized pose representation.
It predicts pose tokens decoded via a learned codebook, providing stronger pose priors and more robust reconstruction.
Despite these design differences, these approaches share a common paradigm that we refer to as the single-view HMR backbone (~\cref{fig:components}(d)).
Such models typically consist of a ViT encoder that extracts image features and a transformer-based learning structure that updates learnable SMPL tokens through attention with the image features. 
The resulting token representation is then mapped to the SMPL parameters $(\theta, \beta)$ and camera coefficients $\Pi$.
\textbf{In our framework}, these pretrained single-view HMR backbones are leveraged as strong priors to enable generalizable multi-view reconstruction without requiring any multi-view training data.
These models provide high-quality initial estimates and impose powerful single-view regularization during test-time optimization, encouraging per-view poses to remain consistent with the image-conditioned latent space learned from large-scale single-view datasets.


\noindent \textbf{2D Keypoint Detector.}
A pretrained and frozen 2D keypoint detector can be employed to extract reliable 2D image-space cues for each view.
Depending on the detector’s training objective, it can provide sparse joint locations, dense correspondences, or mesh-aligned keypoints directly from the input image.
These detections supply strong geometric evidence that is independent of the single-view HMR backbone.
\textbf{In our framework}, these detected 2D keypoints serve as robust constraints on the projection of the 3D joints during test-time optimization, effectively guiding the refinement of pose and shape parameters and improving reconstruction accuracy.

\subsection{Proposed Paradigm}

To generalize to arbitrary camera configurations, we leverage strong priors from pretrained single-view HMR models, then refine them at inference time, therefore eliminating the multi-view training.
Based on this principle, we propose a model-agnostic TTA paradigm for multi-view human mesh recovery. 
Our core approach treats single-view predictions as an initialization, iteratively refining them through cross-view consistency and 2D anatomical cues, while regularization terms prevent the results from drifting excessively from the reliable single-view priors.
In ~\cref{fig:monomv_architecture}(a), a frozen single-view HMR backbone first produces per-view SMPL estimates. 
In addition, we introduce an extra set of optimizable SMPL parameters, termed the \emph{virtual view}. 
The virtual view is initialized from the per-view SMPL predictions.
During TTA, the virtual view parameters are further refined using gradients aggregated from all views. 
After optimization, the virtual view represents the final reconstruction that summarizes the information from all input views. 
Conceptually, the virtual view plays a role similar to aggregating multi-view predictions (e.g., averaging per-view results), but instead of directly fusing predictions, it is optimized through gradients from all views. 
In ~\cref{fig:monomv_architecture}(b), TTA jointly refines the learned SMPL tokens and the virtual-view parameters through 2D observations, multi-view consistency constraints, and regularization terms.

\subsection{Optimizable Component}

\noindent
The choice of the optimizable component \( \Theta \) is crucial for effective test-time adaptation, as the selected component must balance expressiveness, computational efficiency, and alignment with the adaptation objectives.


\noindent \textbf{Single-view Backbone Optimizable Component.}
As illustrated in \cref{fig:components}, there are multiple candidate components within transformer-based single-view HMR models that can be optimized during test-time adaptation within our proposed paradigm, which are broadly categorized into three groups: model parameters (blue), latent tokens (pink and orange), and explicit SMPL parameters (green).
\textbf{Model parameters} encompass the weights of the transformer backbone and decoder head, which encode rich priors learned from large-scale single-view datasets. However, these weights are extremely high-dimensional, making test-time optimization computationally prohibitive and potentially destructive to the pretrained priors.
\textbf{Explicit SMPL parameters} are low-dimensional (72 for pose \(\theta\) and 10 for shape \(\beta\)), making them computationally efficient to optimize. However, optimizing only these parameters fails to exploit the image-conditioned latent structure learned by single-view HMR models. Moreover, the optimization landscape with respect to the SMPL parameters is highly non-linear, which makes the process prone to getting trapped in poor local minima. As a result, the optimization can easily overfit noisy 2D detections or drift toward view-inconsistent solutions.
\textbf{Tokens} provide a compact representation for modeling human pose and shape within transformer-based HMR architectures. Within this category, two common variants can be considered. SMPL query tokens (pink in ~\cref{fig:components}) serve as the initial input tokens to the transformer, lacking sufficiently abstracted and task-specific features to fully support stable multi-view adaptation.
In contrast, learned SMPL tokens (orange in ~\cref{fig:components}) capture richer semantic and image-conditioned representations that are directly instrumental for pose and shape prediction. During optimization of the learned SMPL tokens, gradients are propagated only through the final MLP layers (implemented as a single linear projection) without back-propagating through the transformer backbone. 
\textbf{In our framework}, we therefore adopt learned SMPL tokens as the optimization variables within single-view backbone. This lightweight design avoids full forward and backward passes through the entire network, enabling lightweight TTA while preserving the pretrained backbone priors. Ablation comparisons of different optimization components are provided in Sec.~\ref{sec:ablation}.

\noindent \textbf{Virtual-View Optimizable Component.}
To obtain a unified reconstruction across views, we introduce a set of optimizable SMPL parameters, termed the \textit{virtual view}, which is initialized from per-view predictions and subsequently updated during TTA.
We initialize the virtual-view pose parameters using a weighted averaging strategy over the per-view predictions. Specifically, We compute per-joint statistics across views and apply a $1$-$\delta$ filtering mechanism to suppress unreliable joint rotations.
For each joint rotation $k$, represented using the continuous 6D parameterization~\cite{zhou2019continuity}, we retain the reliable set $\mathcal{S}^{(k)}$:
{\small
\begin{equation}
    \begin{aligned}
        \\
        \mathcal{S}^{(k)} = \left\{ \mathbf{\theta}_i^{(k)} \mid d_i^{(k)} \le \operatorname{Std}(d_1^{(k)}, \dots, d_N^{(k)}) \right\},
        \\ s.t. \quad d_i^{(k)} = \left\lVert \mathbf{\theta}_i^{(k)} - \frac{1}{N} \sum_{i=1}^{N} \mathbf{\theta}_i^{(k)} \right\rVert_2.
    \end{aligned}
\end{equation}}

where $i \in \{1, \dots, N\}$ is the camera view index, and $\mathbf{\theta}_i^{(k)}$ denotes the rotation of joint $k$ from view $i$. We define $i=0$ as the virtual view. The virtual-view pose parameters $\theta_0^{(k)}$ are initialized as the mean of the retained set $\mathcal{S}^{(k)}$.
For the global orientation (i.e., the root joint rotation $\mathbf{\theta}_i^{(0)}$), two settings are considered.
When cameras are calibrated, we first transform the per-view global orientations into the world coordinate frame using the known camera extrinsics, and then apply the weighted averaging described above.
When cameras are uncalibrated (calibration-free mode), global orientations remain view-specific and are not included in the weighted averaging procedure, since they cannot be expressed in a shared coordinate frame. In this case, only the global orientation is view-specific, while all remaining pose rotations are weighted averaged.
Shape parameters are assumed to be view-consistent and initialized as
\(
\boldsymbol{\beta}_0 = \frac{1}{N} \sum_{i=1}^{N} \boldsymbol{\beta}_i.
\)
We denote the virtual-view optimizable SMPL parameters as 
\(
\Theta_0 = (\boldsymbol{\theta}_0, \boldsymbol{\beta}_0).
\)
We empirically observe that this reliability-aware weighted initialization improves stability and final performance compared to simply averaging or canonical pose initialization.
Further analysis of different initialization strategies is provided in Sec.~\ref{sec:ablation}.

\subsection{Test-time Adaptation}
We formalize the TTA paradigm as follows:
\vspace{-10pt}

\begin{align}
    \Theta_{i,t+1} &= \Theta_{i,t} - \eta \nabla_{\Theta_{i,t}}(\mathcal{L}_{\text{2D}}+\lambda \mathcal{L}_{\text{con}}+\gamma \mathcal{L}_{\text{reg}})
      \label{eq:TTA}
\end{align}

\noindent at each optimization step \( t \) during inference, we adaptively update the optimizable component \( \Theta_{i,t} \) under the guidance of three losses: the 2D reprojection loss \( \mathcal{L}_{\text{2D}} \), the cross-view consistency loss \( \mathcal{L}_{\text{con}} \), and the regularization loss \( \mathcal{L}_{\text{reg}} \). The weights \( \lambda, \gamma \) balance the contributions of these losses, and \( \eta \) denotes the learning rate. We will alternatively update the single-view backbone  optimizable component (learned SMPL token) and the virtual view component (SMPL parameters) with their respective losses. For simplicity, the settings of the optimization step, the loss weights, and other hyperparameters are omitted from the loss notation; detailed settings are provided in the Supplementary.

\noindent \textbf{2D Reprojection Loss.}
To provide view-specific supervision during test-time adaptation, we leverage a pretrained DensePose-based anatomical landmark detector~\cite{gozlan2025opencapbench}, which supplies pseudo–ground truth annotations $\hat{P}_i$ for each view. These annotations include 44 joint keypoints (25 OpenPose joints and 19 additional joints) and 35 anatomical surface landmarks, each associated with a confidence score $D_i^{(k)}$.
We incorporate anatomical landmarks because, in contrast to standard joint keypoints, they encode not only spatial positions but also fine-grained joint twist rotation cues~\cite{li2021hybrik}. For example, a wrist twist alters the appearance of the hand surface even though the 3D wrist joint location barely change. Such rotational effects produce significant 2D displacements in the surface landmarks, enabling the supervision to capture subtle pose variations that joint-only signals fail to represent.
To mitigate detector noise, we retain only 2D observations with confidence greater than 0.9. 
For each camera view $i$, the current optimizable component $\Theta_i$ (i.e., the learned SMPL token) is mapped to a 3D mesh $\mathcal{M}_i$ through the MLP and SMPL wrapper, denoted as $\mathcal{M}_i = \mathcal{R}_i(\Theta_i)$.
We then regress 44 joints locations from $\mathcal{M}_i$ using a pretrained keypoint regressor $\mathcal{P}(\cdot)$~\cite{goel2023humans}, and project them to the image plane:
$P_i^{r} = \Pi_i\big( \mathcal{P}(\mathcal{M}_i) \big)$,
where $\Pi_i$ denotes the camera projection function.
The 35 anatomical landmarks $P^a_i$ are obtained directly by indexing the corresponding surface vertices from $\mathcal{M}_i$ via a predefined index list. The 2D reprojection loss for view $i$ is defined as:
\vspace{-3pt}
\begin{equation*}
    \mathcal{L}_{\text{2d}} = \| P^r_i  - \hat{P}^r_i\ \|_2 + \| P^a_i - \hat{P}^a_i \|_2 
\end{equation*}


\noindent \textbf{Cross-view Consistency Loss.}
Cross-view consistency loss is added to enforce agreement of human mesh and geometry between pairs of views $(i, j)$, which is calculated by:
\vspace{-5pt}
\begin{equation*}
    \mathcal{L}_{\text{con}} 
    = \sum^N_{i \neq j} \|\theta_i - \theta_j\|_2 + \|\beta_i - \beta_j\|_2
    + \|\mathcal{M}_i - \mathcal{M}_j\|_2 
\end{equation*}
The above formulation is applied differently depending on the optimized component. 
When updating the virtual-view component, the loss is computed between the virtual view ($i=0$) and each camera view ($j \in \{1,\dots,N\}$), and gradients are applied only to the virtual-view parameters. 
When optimizing the single-view backbone components, the summation is restricted to camera views ($i,j \in \{1,\dots,N\}$), enforcing pairwise agreement among them. In calibration-free mode, the global-orientation cannot be applied, as per-view orientations are not expressed in a common frame. The quadratic complexity of $\mathcal{O}(N^2)$ with respect to the number of views can be reduced to $\mathcal{O}(N)$ by adopting a star-structured consistency graph with negligible performance difference. Further details are provided in the supplementary.

\noindent \textbf{Regularization Loss.}
A regularization loss is also added to prevent large deviations from the initial single-view predictions:
\vspace{-5pt}
\begin{equation*}
    \mathcal{L}_{\text{reg}} 
    = \sum^N_{i=1} \|\theta_i - \theta_{i,0}\|_2 + \|\beta_i - \beta_{i,0}\|_2
    + \|\mathcal{M}_i - \mathcal{M}_{i,0}\|_2
\end{equation*}
where $\theta_{i,0}$, $\beta_{i,0}$, and $\mathcal{M}_{i,0}$ denote the parameters and mesh of view $i$ at iteration $t=0$, i.e., the initial predictions obtained before test-time adaptation.


\vspace{-5pt}
\section{Experiments}

\subsection{Experiment Configuration}
{\bf Single-view Backbones.} We utilize pretrained single-view backbones. For HMR2.0, it's trained with Human3.6M~\cite{human3.6m}, MPI-INF-3DHP~\cite{mono-3dhp2017}, COCO~\cite{lin2014microsoft} and MPII~\cite{andriluka14cvpr}, AVA~\cite{murray2012ava} and AI Challenger~\cite{wu2017ai}. For TokenHMR, it trained the tokenizer with AMASS~\cite{mahmood2019amass} and MOYO~\cite{tripathi2023ipman}. And it follows HMR2.0b's training dataset and includes BEDLAM~\cite{Black_CVPR_2023}. For CameraHMR, it follows hmr2.0 training data except AVA, and adds AGORA and BEDLAM. 

\noindent
{\bf 2D Detectors.}
We choose a pretrain-trained 2D detector, Synthpose~\cite{gozlan2025opencapbench}. It's first pretrained on the original COCO, COCO-WholeBody dataset~\cite{jin2020whole,xu2022zoomnas}, then fine-tuned on the 3DPW and BedLam training dataset. We then adapt the network's head to output 35 anatomical points and 44 skeleton points, and finetune it on the Human3.6M and MPI-INF-3DHP training dataset. Following previous multi-view work~\cite{liang2019shape,shin2020multi}, we train the detector on subjects 1,5,6,7,8 and test on subjects 9, 11 for human3.6m, and train on subjects 1 to 7 and test on subject 8 for MPI-INF-3DHP. 
The reason we fine-tuning is that the SMPL parameters in Human3.6M and MPI-INF-3DHP datasets are pseudo ground truths generated by NeuralAnnot~\cite{moon2022neuralannot}, which exhibits systematic misalignment between the SMPL mesh and the real subjects. 
To ensure consistency between the detected 2D keypoints and the pseudo-GT SMPL annotations used during TTA, we fine-tune the detector on these datasets.

\subsection{Evaluation Metrics}
Following previous works~\cite{liang2019shape,shin2020multi}, we evaluate 3D human mesh recovery on Human3.6M using MPJPE (Mean Per Joint Position Error), PA-MPJPE (Procrustes Analysis Mean Per Joint Position Error), and MPVPE(Mean Per Vertex Position Error).
MPJPE/MPVPE measures the Euclidean distance between predicted and ground truth joints/vertices after aligning the pelvis, and PA means procrustes aligned.
Evaluate on MPI-INF-3DHP using MPJPE, PCK(Percentage of Correct Keypoints), and AUC(Area Under the Curve). PCK means the percentage of keypoints having less than or equal to a predetermined distance from the ground truth. AUC means the area under the PCK curve. 

We evaluate our method using three representative single-view
backbones: CameraHMR~\cite{patel2025camerahmr}, HMR2.0~\cite{goel2023humans}, and TokenHMR~\cite{dwivedi2024tokenhmr}. Unless otherwise stated, we follow a
\(Standard\) setup, which corresponds to our complete multi-view test-time
adaptation pipeline. This standard setup includes:
(1) the 1-delta filtered virtual-view fusion for producing a robust
multi-view initialization,
(2) 2D clues from both keypoints and anatomical landmarks,
(3) multi-view consistency loss together with a regularizer
that anchors the optimization to the initial predictions, and
(4) using the learned token as the evaluation component to be
optimized during TTA. All reported results default to this standard
setup except results in the ablation study.

\subsection{Comparison with State-of-the-art Methods}
Notably, our method is trained solely on single-view data without any multi-view supervision, all the following experiments consistently demonstrate its strong generalization to multi-view camera settings.

\noindent \textbf{Human3.6M.}
Table~\ref{tab:human36m_comparison} reports quantitative results on Human3.6M under the standard protocol.
In calibration setting, our method with CameraHMR backbone achieves 26.9 MPJPE and 20.6 PA-MPJPE, outperforming the previous state of the art, HeatFormer (29.5 MPJPE / 22.4 PA-MPJPE).
Notably, our calibration-free setting achieves substantially lower errors (32.7 MPJPE / 22.0 PA-MPJPE) comparing to Heatformer (42.5 MPJPE / 25.8 PA-MPJPE), highlighting the effectiveness of leveraging strong single-view priors for multi-view refinement without requiring camera calibration.
Among the backbones, CameraHMR backbone exhibits particularly strong robustness in the calibration-free setting. 
We attribute this to its perspective camera formulation and explicit focal length estimation.
These results further validate that a strong single-view prior, combined with TTA, enables effective multi-view reconstruction even without camera calibration.

\begin{table}[t]
\centering
\caption{Comparison on the Human3.6M dataset. Methods are grouped into single-view, multi-view calibration-free, and multi-view calibration-requiring approaches. \textbf{Bold} and \underline{underline} indicate the best and second-best results respectively.}
\label{tab:human36m_comparison}
\scalebox{0.72}{
\begin{tabular}{llcc}
\toprule
 &\textbf{Method} & \textbf{MPJPE} $\downarrow$ & \textbf{PA-MPJPE} $\downarrow$ \\
\midrule
& HMR2.0(a)~\cite{goel2023humans} & 44.8 & 33.6 \\
& HMR2.0(b)~\cite{goel2023humans} & 50.0 & 32.4 \\
Single Camera & HMR2.0+scoreHMR-a~\cite{stathopoulos2024score} & 47.9 & 28.4 \\
& HMR2.0+scoreHMR-b~\cite{stathopoulos2024score} & 44.7 & 29.0 \\
& PostureHMR~\cite{song2024posturehmr} & 44.5 & 31.0 \\
\bottomrule
& Shape Aware~\cite{liang2019shape} & 79.9 & 45.1 \\
& ProHMR~\cite{kolotouros2021probabilistic} & 62.2 & 34.5 \\
& Yu et al~\cite{yu2022multiview} & - & 33.0 \\
& PaFF~\cite{jia2023delving} & 44.8 & 28.2 \\
Calibration-free& HeatFormer~\cite{matsubara2025heatformer} & 42.5 & 25.8 \\
& U-HMR~\cite{li2024human} & \textbf{31.0} & \underline{22.8} \\
& Ours (TokenHMR) & 44.3 & 23.9 \\
& Ours (HMR2.0) & 45.2 & 23.0 \\
& Ours (CameraHMR) & \underline{32.7} & \textbf{22.0} \\
\midrule
& MV-SPIN~\cite{shin2020multi} & 49.8 & 35.4 \\
& PaFF~\cite{jia2023delving} & 33.0 & 26.9 \\
w. Calibration & HeatFormer ~\cite{matsubara2025heatformer} & \underline{29.5} & 22.4 \\
& Ours (TokenHMR) & 32.6 & 23.2 \\
& Ours (HMR2.0) & 31.6 & \underline{21.5} \\
& Ours (CameraHMR) & \textbf{26.9} & \textbf{20.6} \\
\bottomrule
\end{tabular}
}
\end{table}


\noindent
{\bf MPI-INF-3DHP.} The comparison on the MPI-INF-3DHP dataset, as shown in Table~\ref{tab:mpii_comparison}, also highlights our work's performance and generalization. The CameraHMR backbone outperforms previous methods, demonstrating the strong performance and generalizability of our approach.

\begin{table}[t]
\centering
\caption{Comparison results on MPI-INF-3DHP dataset with multi-view methods. \textbf{Bold} indicates the best result, \underline{Underline} indicates the second-best result.}
\scalebox{1}{
\label{tab:mpii_comparison}
\begin{tabular}{lccc}
\toprule
\textbf{Method} & \textbf{MPJPE} $\downarrow$ & \textbf{PCK} $\uparrow$ & \textbf{AUC} $\uparrow$ \\
\midrule
PaFF~\cite{jia2023delving} & 48.4 & 98.6 & 67.3 \\
U-HMR~\cite{li2024human} & \underline{39.7} & 74.0 & \textbf{99.4} \\
HeatFormer~\cite{matsubara2025heatformer} & 39.8 & 99.5 & 72.8 \\
\midrule
Ours (TokenHMR) & 45.2 & \textbf{99.9} & 79.6 \\
Ours (HMR2.0) & 40.3 & \underline{99.6} & 80.7 \\
Ours (CameraHMR) & \textbf{39.0} & \textbf{99.9} & \underline{83.8} \\
\bottomrule
\end{tabular}
}
\end{table}

\noindent
\textbf{Cross-camera Results.} 
Table~\ref{tab:ablation_cross_camera_view} evaluate cross-camera generalization by testing on MPI-INF-3DHP camera set (1, 4, 5, 6), which differ from the original set (0, 2, 7, 8). As shown in Table \ref{tab:ablation_cross_camera_view}, HeatFormer~\cite{matsubara2025heatformer} degrades notably under this camera shift, achieving 45.74 MPJPE. Our method, however, reaches 43.71 MPJPE, 99.44 PCK, and 80.62 AUC, outperforming HeatFormer across all metrics. This demonstrates that our geometry-driven TTA adapts effectively to unseen camera settings, offering strong cross-camera robustness without requiring multi-view training data.


\begin{table}[t]
\centering
\caption{Comparison results on MPI-INF-3DHP dataset with cross camera settings.}
\scalebox{1}{
\label{tab:ablation_cross_camera_view}
\begin{tabular}{cccc}
\toprule
\textbf{Method} & \textbf{MPJPE} $\downarrow$ & \textbf{PCK} $\uparrow$ & \textbf{AUC} $\uparrow$ \\
\midrule
HeatFormer & 45.74 & 99.15 & 68.93 \\
Ours(CameraHMR) & \textbf{43.71} & \textbf{99.44} & \textbf{80.62} \\
\bottomrule
\end{tabular}
}
\end{table}


\subsection{Ablation Study}
\label{sec:ablation}

\noindent
\textbf{Impacts of Virtual-View Initialization and TTA.}
Table~\ref{tab:ablation_virtual_view} shows that the best performance is achieved when both components are enabled. 
Most of the improvement comes from TTA, which substantially refines the initial predictions. 
Without TTA, the initialized virtual view provides only limited gains, indicating that initialization alone cannot resolve multi-view ambiguities but serves as a stable starting point for optimization.

\noindent
{\bf Impacts of Virtual-View Initialization Strategies.}
Table~\ref{tab:ablation_fusion_strategy} compares different initialization strategies for the virtual-view component. 
``No virtual view'' directly uses the averaged per-view estimates after TTA as the final output. 
``T-pose'' initializes the virtual view with a canonical pose, while ``Averaged'' simply averages per-view SMPL parameters. 
``Weighted'' follows the weighted-average strategy in Sec.~3.4.
Across all backbones, introducing a virtual view consistently improves performance, indicating that a unified representation benefits multi-view optimization. Compared to CameraHMR, HMR2.0 and TokenHMR are more sensitive to initialization quality.

\begin{table*}[t]
\centering
\caption{Ablation study on the effects of Virtual View design and TTA.}
\label{tab:ablation_virtual_view}
\scalebox{1}{
\begin{tabular}{cc cccccc}
\toprule
\multicolumn{2}{c}{\textbf{Components}} & \multicolumn{2}{c}{\textbf{Camerahmr}} & \multicolumn{2}{c}{\textbf{HMR2.0}} & \multicolumn{2}{c}{\textbf{TokenHMR}} \\
\cmidrule(lr){1-2} \cmidrule(lr){3-4} \cmidrule(lr){5-6} \cmidrule(lr){7-8}
Virtual View & TTA & \textbf{MPJPE} $\downarrow$ & \textbf{PA-MPJPE} $\downarrow$ & \textbf{MPJPE} $\downarrow$ & \textbf{PA-MPJPE} $\downarrow$ & \textbf{MPJPE} $\downarrow$ & \textbf{PA-MPJPE} $\downarrow$ \\
\midrule
\ding{55} & \ding{51} & \underline{27.8} & \underline{22.7} & \underline{35.1} & 24.1 & 43.9 & 27.5 \\
\ding{51} & \ding{55} & 44.2 & 29.9 & 44.4 & \underline{24.0} & \underline{37.1} & \underline{25.7} \\
\ding{51} & \ding{51} & \textbf{26.9} & \textbf{20.6} & \textbf{31.6} & \textbf{21.5} & \textbf{32.6} & \textbf{23.2} \\
\bottomrule
\end{tabular}
}
\end{table*}

\begin{table*}[t!]
\centering
\caption{Ablation study on virtual view initialization strategy.}
\label{tab:ablation_fusion_strategy}
\scalebox{1}{
\begin{tabular}{lcccccc}
\toprule
& \multicolumn{2}{c}{\textbf{Camerahmr}} & \multicolumn{2}{c}{\textbf{HMR2.0}} & \multicolumn{2}{c}{\textbf{TokenHMR}} \\
\cmidrule(lr){2-3} \cmidrule(lr){4-5} \cmidrule(lr){6-7}
\textbf{Strategy} & \textbf{MPJPE} $\downarrow$ & \textbf{PA-MPJPE} $\downarrow$ & \textbf{MPJPE} $\downarrow$ & \textbf{PA-MPJPE} $\downarrow$ & \textbf{MPJPE} $\downarrow$ & \textbf{PA-MPJPE} $\downarrow$ \\
\midrule
No Virtual View & \underline{27.8} & \underline{22.7} & \underline{35.1} & 24.1 & 43.9 & 27.5 \\
T-pose & 27.2 & 21.2 & 32.5 & 22.4 & 33.8 & 23.6 \\
Averaged & 27.1 & 20.7 & 32.1 & 21.7 & 33.5 & 23.3 \\
Weighted & \textbf{26.9} & \textbf{20.6} & \textbf{31.6} & \textbf{21.5} & \textbf{32.6} & \textbf{23.2} \\
\bottomrule
\end{tabular}
}
\end{table*}

\noindent
{\bf Impacts of TTA Loss Components.}
We isolate the effect of different guidance signals used
during TTA. 
We consider five types of settings: (1) only 2D anatomical landmarks, (2) only 2D
keypoints, (3) their combination, (4) an additional multi-view
consistency, and (5) The standard setting further includes a regularization term.
Reported in Table~\ref{tab:ablation_tta_guidance}, using either anatomical
landmarks or keypoints alone provides moderate improvements, while combining
them yields stronger 2D supervision. The largest performance gain comes from
introducing the multi-view consistency term, which effectively exploits
geometric relationships across views. Adding the regularization term delivering the best and most stable results across all three backbones.

\begin{table*}[t!]
\centering
\caption{Ablation study on TTA loss components.}
\label{tab:ablation_tta_guidance}
\scalebox{1}{
\begin{tabular}{lcccccc}
\toprule
& \multicolumn{2}{c}{\textbf{Camerahmr}} & \multicolumn{2}{c}{\textbf{HMR2.0}} & \multicolumn{2}{c}{\textbf{TokenHMR}} \\
\cmidrule(lr){2-3} \cmidrule(lr){4-5} \cmidrule(lr){6-7}
\textbf{Guidance} & \textbf{MPJPE} $\downarrow$ & \textbf{PA-MPJPE} $\downarrow$ & \textbf{MPJPE} $\downarrow$ & \textbf{PA-MPJPE} $\downarrow$ & \textbf{MPJPE} $\downarrow$ & \textbf{PA-MPJPE} $\downarrow$ \\
\midrule
anatomical (ana) & 39.9 & 29.4 & 35.2 & 24.6 & 33.5 & 24.5 \\
keypoint (kp) & 39.6 & 28.5 & 35.0 & 24.0 & 33.2 & 23.7 \\
kp+ana & 36.8 & 27.2 & 32.3 & 22.5 & 33.0 & 23.6 \\
kp+ana+consistency (con) & \underline{27.3} & \underline{20.9} & \underline{32.2} & \underline{21.6} & \underline{32.9} & \underline{23.4} \\
kp+ana+con+regularizer & \textbf{26.9} & \textbf{20.6} & \textbf{31.6} & \textbf{21.5} & \textbf{32.6} & \textbf{23.2} \\
\bottomrule
\end{tabular}
}
\end{table*}

\begin{table*}[t!]
\centering
\caption{Ablation study on single-view backbone optimizable component.}
\label{tab:ablation_tta_component}
\scalebox{1}{
\begin{tabular}{lcccccc}
\toprule
& \multicolumn{2}{c}{\textbf{Camerahmr}} & \multicolumn{2}{c}{\textbf{HMR2.0}} & \multicolumn{2}{c}{\textbf{TokenHMR}} \\
\cmidrule(lr){2-3} \cmidrule(lr){4-5} \cmidrule(lr){6-7}
\textbf{Component} & \textbf{MPJPE} $\downarrow$ & \textbf{PA-MPJPE} $\downarrow$ & \textbf{MPJPE} $\downarrow$ & \textbf{PA-MPJPE} $\downarrow$ & \textbf{MPJPE} $\downarrow$ & \textbf{PA-MPJPE} $\downarrow$ \\
\midrule
SMPL Parameters & 58.2 & 32.2 & 53.0 & 31.9 & 48.4 & 31.1 \\
Decoder  & 30.8 & 22.8 & 37.2 & 23.3 & 36.9 & 27.5 \\
Learnable Tokens & \underline{29.7} & \underline{20.9} & \underline{36.0} & \underline{23.0} & \underline{35.2} & \underline{25.7} \\
Learned Tokens & \textbf{26.9} & \textbf{20.6} & \textbf{31.6} & \textbf{21.5} & \textbf{32.6} & \textbf{23.2} \\
\bottomrule
\end{tabular}
}
\end{table*}

\noindent
{\bf Impacts of Single-view Backbone Optimizable Component.}
As shown in \cref{tab:ablation_tta_component}, optimizing the learned SMPL tokens achieves the best performance across all backbones, yielding the lowest MPJPE and PA-MPJPE. 
We hypothesize that these tokens capture rich image-conditioned representations that are directly used for pose and shape prediction. In contrast, optimizing the input tokens or decoder parameters is less effective, as the former lacks sufficient task-specific semantics and the latter is less amenable to refinement from a single test sample. Directly optimizing SMPL parameters performs the worst due to their limited dimensionality, and susceptibility to noisy 2D cues.


\begin{table*}[t!]
\centering
\caption{Ablation study on Virtual-view Optimizable Component.}
\label{tab:ablation_consistent_component}
\scalebox{1}{
\begin{tabular}{lcccccc}
\toprule
& \multicolumn{2}{c}{\textbf{Camerahmr}} & \multicolumn{2}{c}{\textbf{HMR2.0}} & \multicolumn{2}{c}{\textbf{TokenHMR}} \\
\cmidrule(lr){2-3} \cmidrule(lr){4-5} \cmidrule(lr){6-7}
\textbf{Component} & \textbf{MPJPE} $\downarrow$ & \textbf{PA-MPJPE} $\downarrow$ & \textbf{MPJPE} $\downarrow$ & \textbf{PA-MPJPE} $\downarrow$ & \textbf{MPJPE} $\downarrow$ & \textbf{PA-MPJPE} $\downarrow$ \\
\midrule
Pose & 31.6 & 21.9 & 35.2 & 24.6 & 34.6 & 23.3 \\
Pose+Orientation & \underline{29.0} & \underline{21.4} & \underline{33.4} & \underline{22.0} & \underline{34.4} & \underline{23.3} \\
Pose+Orientation+Shape & \textbf{26.9} & \textbf{20.6} & \textbf{31.6} & \textbf{21.5} & \textbf{32.6} & \textbf{23.2} \\
\bottomrule
\end{tabular}
}
\end{table*}

\noindent
{\bf Impacts of Virtual-view Optimizable Component.}
Ablations on optimizing virtual-view components shown in \cref{tab:ablation_consistent_component} demonstrate that jointly adapting pose, global orientation, and shape consistently achieves the lowest MPJPE and PA-MPJPE across all backbones. Updating pose only leads to suboptimal performance due to unresolved orientation and shape mismatches, while adding orientation already yields clear gains. Incorporating shape further stabilizes cross-view alignment. While CameraHMR and HMR2.0 exhibit large gains from adding orientation and shape, TokenHMR shows smaller improvements, likely due to its discrete pose prior constraining updates.


\subsection{Qualitative Result}

\begin{figure*}[t]
    \centering
    \includegraphics[width=1\linewidth]{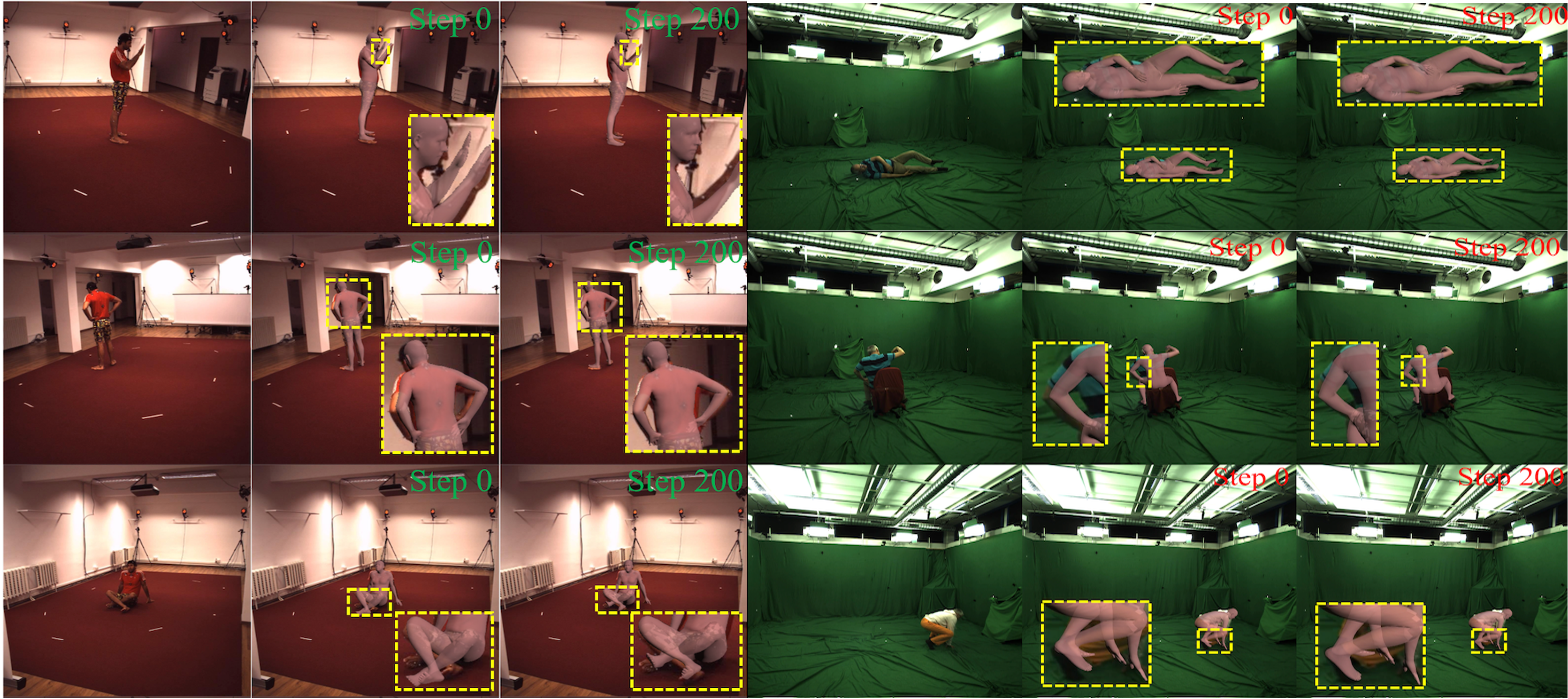}
    \caption{Mesh alignment on the Human3.6M~\cite{human3.6m} (left) and MPI-INF-3DHP~\cite{mono-3dhp2017} (right). From left to right: input image, mesh at Step 0, and mesh at step 200.}
    \label{fig:qualitative_human3.6_3dhp}
\end{figure*}





\begin{figure*}[!t]
    \centering
    \includegraphics[width=1\linewidth]{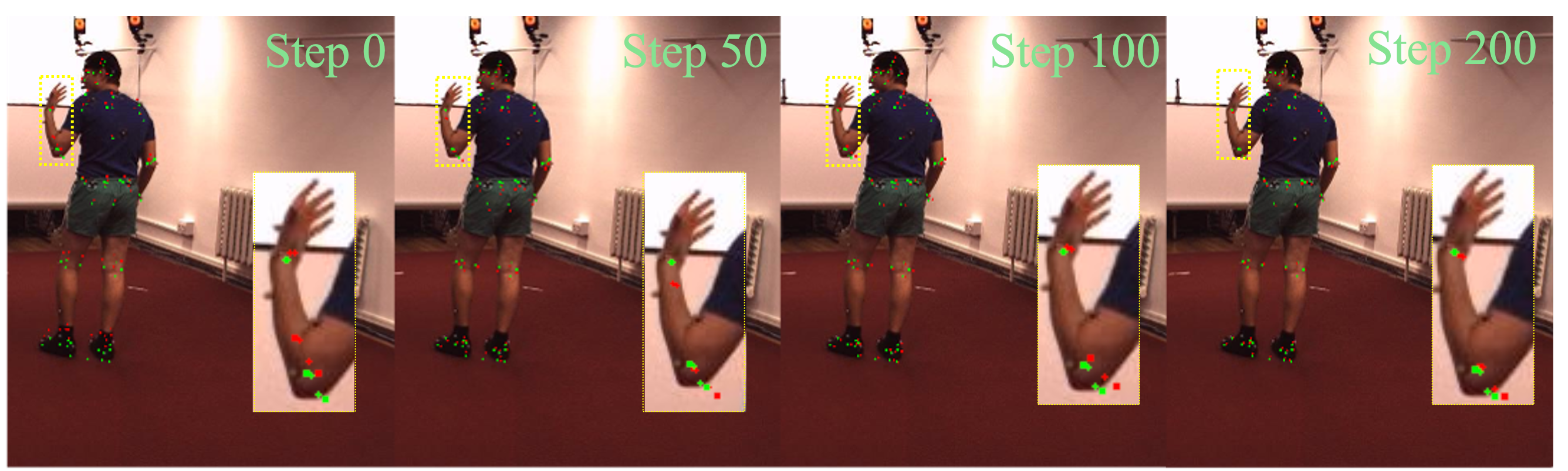}
    \caption{TTA 2D clues alignment. Green markers denote the pseudo ground-truth from 2D detector, while the red markers represent the model’s predictions.}
    \label{fig:tta}
\end{figure*}

\cref{fig:qualitative_human3.6_3dhp}
shows the predicted SMPL mesh on Human3.6M and MPI-INF-3DHP. At Step~0, the initial single-view predictions often exhibit noticeable errors. After 200 TTA steps, these errors are consistently corrected. \cref{fig:tta} visualizes the progression of the test-time adaptation. As the TTA steps increase, the predicted 2D joints gradually converge toward the pseudo-GT keypoints. More qualitative results are presented in Supplementary.

\section{Conclusion}
In this paper, we present a training-free framework for robust and accurate multi-view human mesh recovery. 
By leveraging pretrained single-view HMR models as strong priors, our method introduces a virtual-view representation that is iteratively refined via test-time adaptation guided by 2D anatomical landmarks and multi-view consistency. 
Extensive experiments demonstrate that our approach achieves state-of-the-art performance on benchmark datasets and generalizes effectively to unseen multi-view camera configurations without requiring any multi-view training data.

\section*{Acknowledgments}
This work was supported in part by the US National Science Foundation under Grants IIS-2348427 and 2008447.
{
    \small
    \bibliographystyle{ieeenat_fullname}
    \bibliography{main}
}
\clearpage
\setcounter{page}{1}
\maketitlesupplementary

\section{Overview}

This supplementary material provides additional implementation details, extended ablation studies, and qualitative results that complement the main paper. The contents are organized as follows.

\begin{itemize}
    \item \textbf{Sec.~2: Implementation Details.}
    We provide detailed descriptions of the optimization settings used in TTA, including hyper-parameter configurations, the number of optimization steps, gradient clipping, and warm-up steps. We also report additional measurements of the computational efficiency of our approach, including inference time and optimization cost.
    
    \item \textbf{Sec.~3: Additional Ablation Studies.}
    We present additional quantitative analyses to further understand our proposed framework. These include the impacts of the number of input views, the number of TTA steps, 2D detector quality, learning rate, and etc.

    \item \textbf{Sec.~4: Additional Qualitative Results.}
    We provide additional qualitative visualizations on challenging scenes, including \textit{Block} and \textit{Shelf}, to demonstrate the generalization to unseen camera settings and the robustness under occlusions.

\end{itemize}

\section{Implementation Details}

\noindent {\bf Hyper-parameters.}
The weights of the loss terms (omitted in the main paper for brevity) are included here in the full loss formulation. For the single-view optimizable component, the following losses are used.

\begin{equation*}
    \mathcal{L}_{\text{2d}} = \lambda_{kp}\| P^r_i  - \hat{P}^r_i\ \|_2 + \lambda_{ana}\| P^a_i - \hat{P}^a_i \|_2 
\end{equation*}

{\small
\begin{equation*}
\begin{aligned}
    \mathcal{L}_{\text{con}} 
    = \sum^N_{i \neq j} \lambda_{con\_orient}\|\mathbf{\theta}_i^{(0)} - \mathbf{\theta}_j^{(0)}\|_2 + \lambda_{con\_pose}\|\mathbf{\theta}_i^{(k)} - \mathbf{\theta}_j^{(k)}\|_2 
    \\ + \lambda_{con\_beta}\|\beta_i - \beta_j\|_2
    + \lambda_{con\_vertice}\|\mathcal{M}_i - \mathcal{M}_j\|_2 
\end{aligned}
\end{equation*}}

{\small
\begin{equation*}
\begin{aligned}
    \mathcal{L}_{\text{reg}} 
    = \sum^N_{i=1} \lambda_{reg\_orient}\|\mathbf{\theta}_i^{(0)} - \mathbf{\theta}_{i,0}^{(0)}\|_2
    + \lambda_{reg\_pose}\|\theta_i^{(k)} - \mathbf{\theta}_{i,0}^{(k)}\|_2
    \\+ \lambda_{reg\_beta}\|\beta_i - \beta_{i,0}\|_2
    + \lambda_{reg\_vertice}\|\mathcal{M}_i - \mathcal{M}_{i,0}\|_2
\end{aligned}
\end{equation*}}

When updating the virtual-view component, the consistency loss $\mathcal{L}_{\text{virtual\_con}}$ is computed between the virtual view (denoted as $0$) and each camera view
$j \in \{1,\dots,N\}$:

{\small
\begin{equation*}
\begin{aligned}
\mathcal{L}_{\text{virtual\_con}}
= \sum_{j=1}^{N} &\lambda_{\text{virtual\_orient}}
\|\mathbf{\theta}_0^{(0)} - \mathbf{\theta}_j^{(0)}\|_2 + \lambda_{\text{virtual\_pose}}
\|\mathbf{\theta}_0^{(k)} - \mathbf{\theta}_j^{(k)}\|_2 \\
&+ \lambda_{\text{virtual\_beta}}
\|\beta_0 - \beta_j\|_2 + \lambda_{\text{virtual\_vertice}}
\|\mathcal{M}_0 - \mathcal{M}_j\|_2
\end{aligned}
\end{equation*}}

The hyper-parameters used for different models are summarized in Table~\ref{tab:Hyper-parameters}. All hyper-parameters are initially tuned based on the CameraHMR~\cite{patel2025camerahmr} backbone, and then slightly adjusted for HMR2.0~\cite{goel2023humans} and TokenHMR~\cite{dwivedi2024tokenhmr}. In all experiments, the number of TTA steps is fixed to 200. We also employ a warm-up stage of 20 steps and apply gradient clipping with a norm of 0.1 to stabilize the optimization.

\begin{table}
\centering
\caption{Hyper-parameters for each model.}
\label{tab:Hyper-parameters}
\scalebox{0.9}{
\begin{tabular}{lccc}
\toprule
Hyper-parameter & {\textbf{Camerahmr}} & {\textbf{HMR2.0}} & {\textbf{TokenHMR}} \\
\midrule
$\eta$ & $6e^{-2}$ & $1e^{-1}$ & $6e^{-2}$ \\
$\lambda_{kp}$ & $3e^{-1}$ & $3e^{-1}$ & $3e^{-1}$ \\
$\lambda_{ana}$ & $3e^{-1}$ & $3e^{-1}$ & $3e^{-1}$ \\
$\lambda_{reg\_orient}$ & $3e^{1}$ & $1e^{1}$ & $1e^{1}$ \\
$\lambda_{reg\_pose}$ & $1e^{-1}$ & $1e^{-1}$ & $1e^{-1}$ \\
$\lambda_{reg\_betas}$ & $2e^{-2}$ & $2e^{-2}$ & $2e^{-2}$ \\
$\lambda_{reg\_vertice}$ & $1e^{-2}$ & $1e^{-2}$ & $1e^{-2}$\\
$\lambda_{con\_orient}$ & $5e^{0}$ & $5e^{0}$ & $1e^{0}$ \\
$\lambda_{con\_pose}$ & $5e^{0}$ & $5e^{0}$ & $1e^{0}$ \\
$\lambda_{con\_betas}$ & $5e^{0}$ & $5e^{0}$ & $1e^{0}$ \\
$\lambda_{con\_vertice}$ & $3e^{-1}$ & $3e^{-1}$ & $1.5e^{-1}$ \\
\midrule
$\eta_{virtual}$ & $1e^{-2}$ & $1e^{-2}$ & $1e^{-2}$\\
$\lambda_{virtual\_orient}$ & $3e^2$ & $3e^2$ & $3e^2$ \\
$\lambda_{virtual\_pose}$ & $1e^1$ & $1e^1$ & $1e^1$ \\
$\lambda_{virtual\_vertice}$ & $1e^{-1}$ & $1e^{-1}$ & $1e^{-1}$ \\
$\lambda_{virtual\_betas}$ & $1e^{0}$ & $1e^{0}$ & $1e^{0}$ \\
\midrule
\bottomrule
\end{tabular}}
\end{table}


\noindent {\bf Computational Efficiency.}
The original formulation of the cross-view consistency loss considers all pairwise relations between camera views, which requires evaluating all view pairs and therefore has a computational complexity of $\mathcal{O}(N^2)$.

To reduce the computational cost, we instead adopt a star-structured consistency graph. Specifically, we introduce a reference representation computed as the arithmetic mean of the SMPL parameters across all views:
\[
\bar{\theta} = \frac{1}{N}\sum_{i=1}^{N}\theta_i\ ,
\qquad
\bar{\beta} = \frac{1}{N}\sum_{i=1}^{N}\beta_i\ ,
\qquad
\bar{\mathcal{M}} = \frac{1}{N}\sum_{i=1}^{N}\mathcal{M}_i\ .
\]

The consistency loss is then defined as the deviation of each view from this reference point:
\[
\mathcal{L}_{\text{con\_star}} =
\sum_{i=1}^{N}
\left(
\|\theta_i - \bar{\theta}\|_2 +
\|\beta_i - \bar{\beta}\|_2 +
\|\mathcal{M}_i - \bar{\mathcal{M}}\|_2
\right).
\]

Under this formulation, each view only interacts with the reference node, resulting in a star-shaped graph structure. The number of constraints therefore scales linearly with the number of views, reducing the complexity from $\mathcal{O}(N^2)$ to $\mathcal{O}(N)$.

Table~\ref{tab:re_ablation_tta_steps} reports the accuracy and inference time under different TTA steps using the CameraHMR backbone on the Human3.6M dataset, achieving an Average Inference Time per Image (AITI) of 2.5 seconds with 200 TTA steps on a single mid-grade GPU (NVIDIA RTX A5000).. With the star-structured consistency graph, the final performance slightly decreases from 26.9 MPJPE / 20.6 PA-MPJPE (fully connected formulation) to 27.56 MPJPE / 21.61 PA-MPJPE.

\begin{table}
\centering
\caption{Inference time with star-structured consisntecy graph.}
\label{tab:re_ablation_tta_steps}
\scalebox{1}{
\begin{tabular}{r ccc}
\toprule
\textbf{Steps} & \textbf{MPJPE}↓ & \textbf{PA-MPJPE}↓ & \textbf{AITI(s)}↓ \\
\midrule
0 & 44.23 & 29.86 & 0.03\\
50 & 36.54 & 26.64 & 0.63\\
100 & 30.36 & 23.26 & 1.26\\
150 & 28.52 & 22.17 & 1.88\\
200 & 27.56 & 21.61 & 2.50 \\
\bottomrule
\end{tabular}}
\end{table}


\section{Additional Ablation Studies}


\textbf{Impact of View Number.} 
Table~\ref{tab:ablation_number_of_views} evaluates the effect of using different numbers of input views during test-time adaptation on Human3.6m~\cite{human3.6m} when the CameraHMR backbone is used. More views consistently improve reconstruction accuracy: moving from 2 to 3 views yields a clear reduction in both MPJPE and PA-MPJPE, and using 4 views achieves the best performance by providing stronger geometric constraints and more reliable multi-view consensus. These results highlight that additional views substantially enhance the effectiveness of our TTA framework by supplying richer cross-view supervision.


\begin{table}
\centering
\caption{Ablation study on the number of views.}
\scalebox{1}{
\label{tab:ablation_number_of_views}
\begin{tabular}{cccc}
\toprule
\textbf{\# of view} & \textbf{MPJPE} $\downarrow$ & \textbf{PA-MPJPE} $\downarrow$ \\
\midrule
2 & 31.2 & 23.0 \\
3 & \underline{29.5} & \underline{22.0} \\
4 & \textbf{26.9} & \textbf{20.6} \\
\bottomrule
\end{tabular}
}
\end{table}


\noindent \textbf{Impact of Learning Rate.}
Table~\ref{tab:ablation_learning_rate} reports the learning-rate ablation using the CameraHMR backbone on the Human3.6M dataset. In this experiment, the number of TTA steps is fixed to 200 for all settings. Increasing the learning rate from $1\times10^{-2}$ to $1\times10^{-1}$ consistently improves both MPJPE and PA-MPJPE, with $6\times10^{-2}$ yielding the best performance. A larger learning rate such as $1\times10^{-1}$ begins to destabilize the optimization. These results indicate that moderately large updates are beneficial for effective test-time adaptation, whereas overly aggressive updates degrade performance. Therefore, we adopt $6\times10^{-2}$ as the default learning rate for the CameraHMR backbone.


\begin{table}
\centering
\caption{Ablation study on learning rate.}
\scalebox{1}{
\label{tab:ablation_learning_rate}
\begin{tabular}{cccc}
\toprule
\textbf{learning rate} & \textbf{MPJPE} $\downarrow$ & \textbf{PA-MPJPE} $\downarrow$ \\
\midrule
$1e^{-2}$ & 29.8 & 21.1 \\
$3e^{-2}$ & \underline{27.4} & \underline{21.0} \\
$6e^{-2}$ & \textbf{26.9} & \textbf{20.6} \\
$1e^{-1}$ & 29.5 & 22.0 \\
\bottomrule
\end{tabular}
}
\end{table}


\noindent\textbf{Impact of TTA Steps.}
Based on empirical observations, we set the number of TTA optimization steps to 200. Although fewer steps can still achieve good performance with carefully tuned hyper-parameters (e.g., using a larger learning rate), such configurations tend to be model-specific and do not generalize well across different single-view backbones. Instead, we increase the number of optimization steps while reducing the learning rate and adjusting other hyper-parameters accordingly, which leads to a more stable optimization process and better cross-model generalization.

Table~\ref{tab:ablation_tta_steps} reports the impact of TTA steps. CameraHMR and TokenHMR continue to benefit from additional optimization steps, with the best results obtained at 200 steps. For the HMR2.0 backbone, strong performance is already achieved with only 100 optimization steps, and the best performance is reached at 150 steps. Beyond this point, the results remain very stable with negligible variation. Considering both performance and generalization across different backbones, we adopt 200 TTA steps as the default setting in all experiments.

\begin{table*}[!t]
\centering
\caption{Ablation study on TTA steps.}
\label{tab:ablation_tta_steps}
\scalebox{1}{
\begin{tabular}{rccccccc}
\toprule
\multicolumn{1}{c}{\textbf{}}& \multicolumn{2}{c}{\textbf{Camerahmr}} & \multicolumn{2}{c}{\textbf{HMR2.0}} & \multicolumn{2}{c}{\textbf{TokenHMR}} \\
\cmidrule(lr){2-3} \cmidrule(lr){4-5} \cmidrule(lr){6-7}
\textbf{Steps} & \textbf{MPJPE} $\downarrow$ & \textbf{PA-MPJPE} $\downarrow$ & \textbf{MPJPE} $\downarrow$ & \textbf{PA-MPJPE} $\downarrow$ & \textbf{MPJPE} $\downarrow$ & \textbf{PA-MPJPE} $\downarrow$ \\
\midrule
50  & 35.6 & 25.9 & 35.4 & \underline{22.0} & 35.1 & 24.3 \\
100 & 29.9 & 22.3 & 32.4 & \textbf{21.5} & 33.5 & 23.6 \\
150 & 28.0 & 21.2 & \textbf{31.6} & \textbf{21.5} & 32.8 & \underline{23.3} \\
200 & \textbf{26.9} & \textbf{20.6} & \textbf{31.6} & \textbf{21.5} & \textbf{32.6} & \textbf{23.2} \\
250 & \underline{27.4} & \underline{21.0} & \underline{31.7} & \textbf{21.5} & \underline{32.7} & \underline{23.3} \\
\bottomrule
\end{tabular}
}
\end{table*}


\noindent \textbf{Impact of 2D Detector Quality.} Table ~\ref{tab:ablation_2d_detector} examines how the quality of the 2D detector influences TTA performance on Human3.6m evaluation dataset. As the detector becomes more accurate (lower end-point-error, EPE), all backbones show consistent improvements in MPJPE and PA-MPJPE. Finetuning the detector with 1k/2k/full-evaluation data steadily reduces 2D error and yields clear performance gains. TTA with ground-truth 2D clues achieves the best results, revealing the upper bound of our framework. Notably, CameraHMR benefits the most, likely because its full-perspective camera model provides better geometric alignment between 2D cues and 3D predictions; in contrast, the weak-perspective models in HMR2.0 and TokenHMR limit how much accurate 2D supervision can help. Overall, these results show that TTA is highly sensitive to the reliability of 2D cues, and that stronger detectors yield significantly better refinement, revealing substantial headroom for further improvement.


\begin{table*}[!t]
\centering
\caption{Ablation study on quality of 2D detector.}
\label{tab:ablation_2d_detector}
\scalebox{0.95}{
\begin{tabular}{lccccccc}
\toprule
\multicolumn{2}{c}{\textbf{}}& \multicolumn{2}{c}{\textbf{CameraHmr}} & \multicolumn{2}{c}{\textbf{HMR2.0}} & \multicolumn{2}{c}{\textbf{TokenHMR}} \\
\cmidrule(lr){3-4} \cmidrule(lr){5-6} \cmidrule(lr){7-8}
\textbf{2D detector} & \textbf{EPE} & \textbf{MPJPE} $\downarrow$ & \textbf{PA-MPJPE} $\downarrow$ & \textbf{MPJPE} $\downarrow$ & \textbf{PA-MPJPE} $\downarrow$ & \textbf{MPJPE} $\downarrow$ & \textbf{PA-MPJPE} $\downarrow$ \\
\midrule
w/o evaluation data(ours) & 5.94 & 26.9 & 20.6 & 31.6 & 21.5 & 32.6 & 23.2 \\
\midrule
w/ 1k evaluation data & 5.05 & 23.8 & 18.0 & 30.1 & 20.3 & 31.7 & 22.2 \\
w/ 2k evaluation data & 4.74 & 21.8 & 16.1 & 29.9 & 20.1 & 31.5 & 21.6 \\
w/ whole evaluation data & \underline{2.41} & \underline{19.0} & \underline{13.2} & \underline{27.6} & \underline{18.3} & \underline{31.0} & \underline{21.4} \\
ground truth & \textbf{0} & \textbf{15.8} & \textbf{10.4} & \textbf{26.7} & \textbf{17.5} & \textbf{30.0} & \textbf{20.7} \\
\bottomrule
\end{tabular}
}
\end{table*}


\noindent\textbf{Impact of Gradient Clipping.}
To stabilize the optimization during test-time adaptation, we employ gradient clipping. Specifically, the gradient norm is clipped to $0.1$ to prevent excessively large parameter updates that may destabilize the optimization. An ablation study of gradient clipping is provided in Table~\ref{tab:ablation_clip}.

\begin{table*}[t]
\centering
\caption{Ablation study on gradient clip norm during TTA}
\label{tab:ablation_clip}
\scalebox{1}{
\begin{tabular}{lcccccc}
\toprule
& \multicolumn{2}{c}{\textbf{Camerahmr}} & \multicolumn{2}{c}{\textbf{HMR2.0}} & \multicolumn{2}{c}{\textbf{TokenHMR}} \\
\cmidrule(lr){2-3} \cmidrule(lr){4-5} \cmidrule(lr){6-7}
\textbf{TTA Strategy} & \textbf{MPJPE} $\downarrow$ & \textbf{PA-MPJPE} $\downarrow$ & \textbf{MPJPE} $\downarrow$ & \textbf{PA-MPJPE} $\downarrow$ & \textbf{MPJPE} $\downarrow$ & \textbf{PA-MPJPE} $\downarrow$ \\
\midrule
w/o gradient clip norm & 30.5 & 22.5 & 37.0 & 24.6 & 35.2 & 24.1 \\
w/ gradient clip norm & \textbf{26.9} & \textbf{20.6} & \textbf{31.6} & \textbf{21.5} & \textbf{32.6} & \textbf{23.2} \\
\bottomrule
\end{tabular}
}
\end{table*}

\section{Additional Qualitative Results}

\begin{figure*}[!t]
    \centering
    \includegraphics[width=0.9\linewidth]{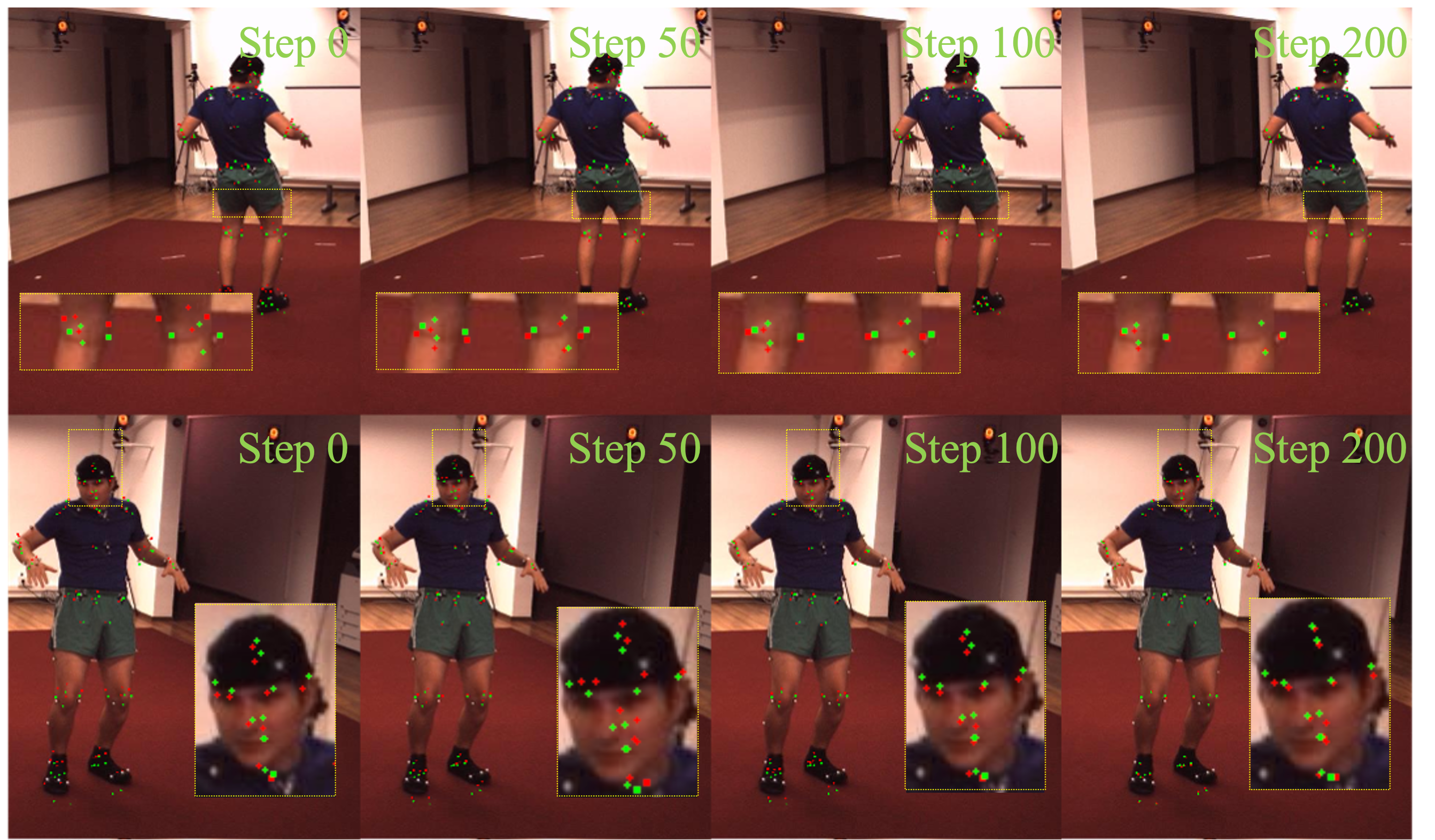}
    \caption{TTA 2D clues alignment. Green markers denote the pseudo ground-truth from 2D detector, while the red markers represent the model’s predictions.}
    \label{fig:sup_tta_h36m}
\end{figure*}

\cref{fig:sup_tta_h36m} visualizes the progression of the TTA process. As the number of TTA steps increases, the predicted 2D joints gradually converge toward the pseudo-GT keypoints provided by the 2D detector, demonstrating the effectiveness of our optimization process.

\begin{figure*}[!t]
    \centering
    \includegraphics[width=1\linewidth]{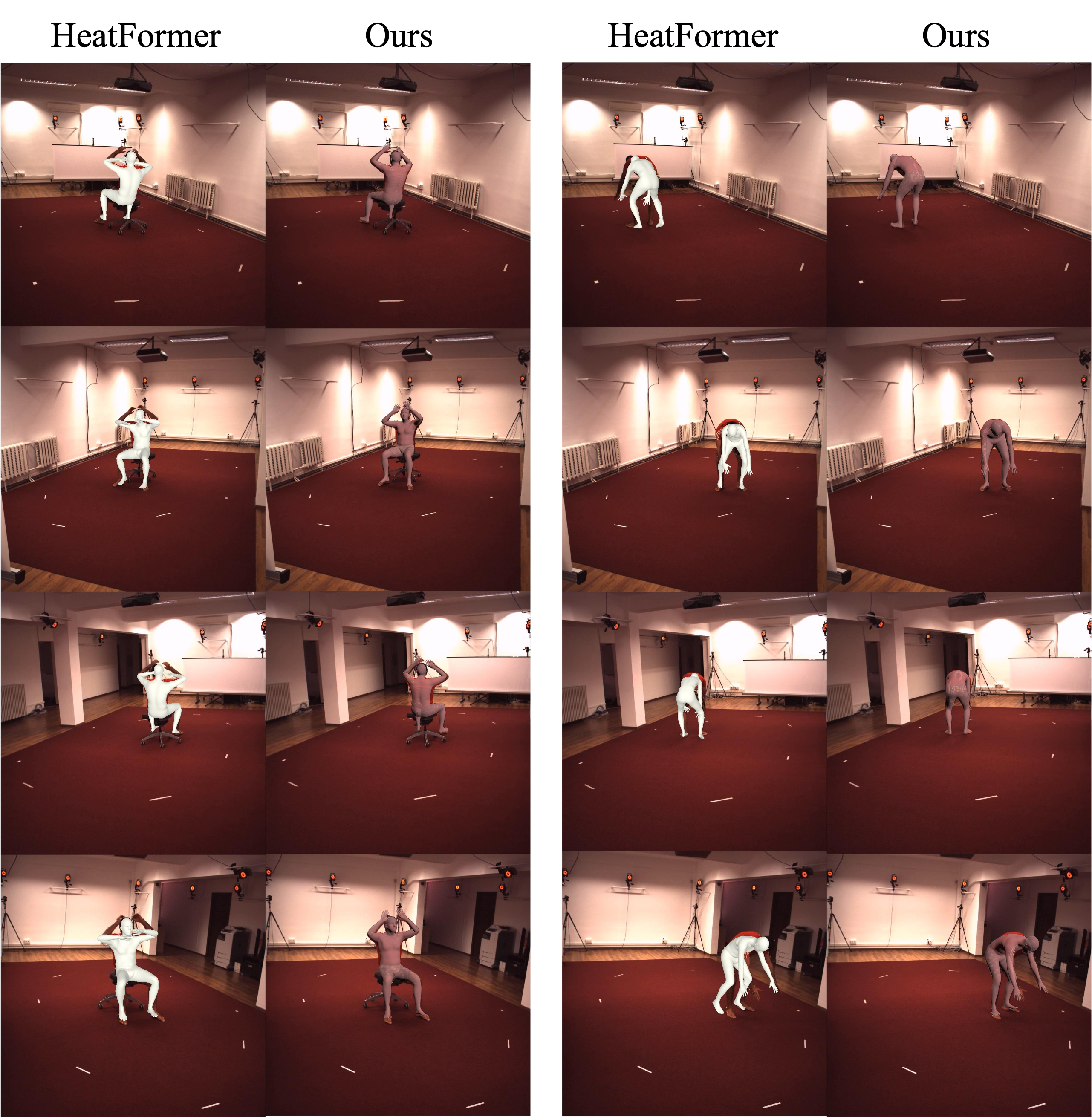}
    \caption{Qualitative comparison between HeatFormer and our method on the Human3.6M dataset. Pink meshes denote our predictions, while white meshes correspond to HeatFormer (iteration 4).}
    \label{fig:Sup_compare_1}
\end{figure*}

\cref{fig:Sup_compare_1} presents a qualitative comparison with HeatFormer on the Human3.6M dataset. The HeatFormer results are generated using the iteration-4 pretrained checkpoint from their public code. In the visualization, the pink meshes correspond to our method, while the white meshes represent the predictions from HeatFormer. Our method produces more accurate mesh alignment with the underlying human pose.

\begin{figure*}[!t]
    \centering
    \includegraphics[width=1\linewidth]{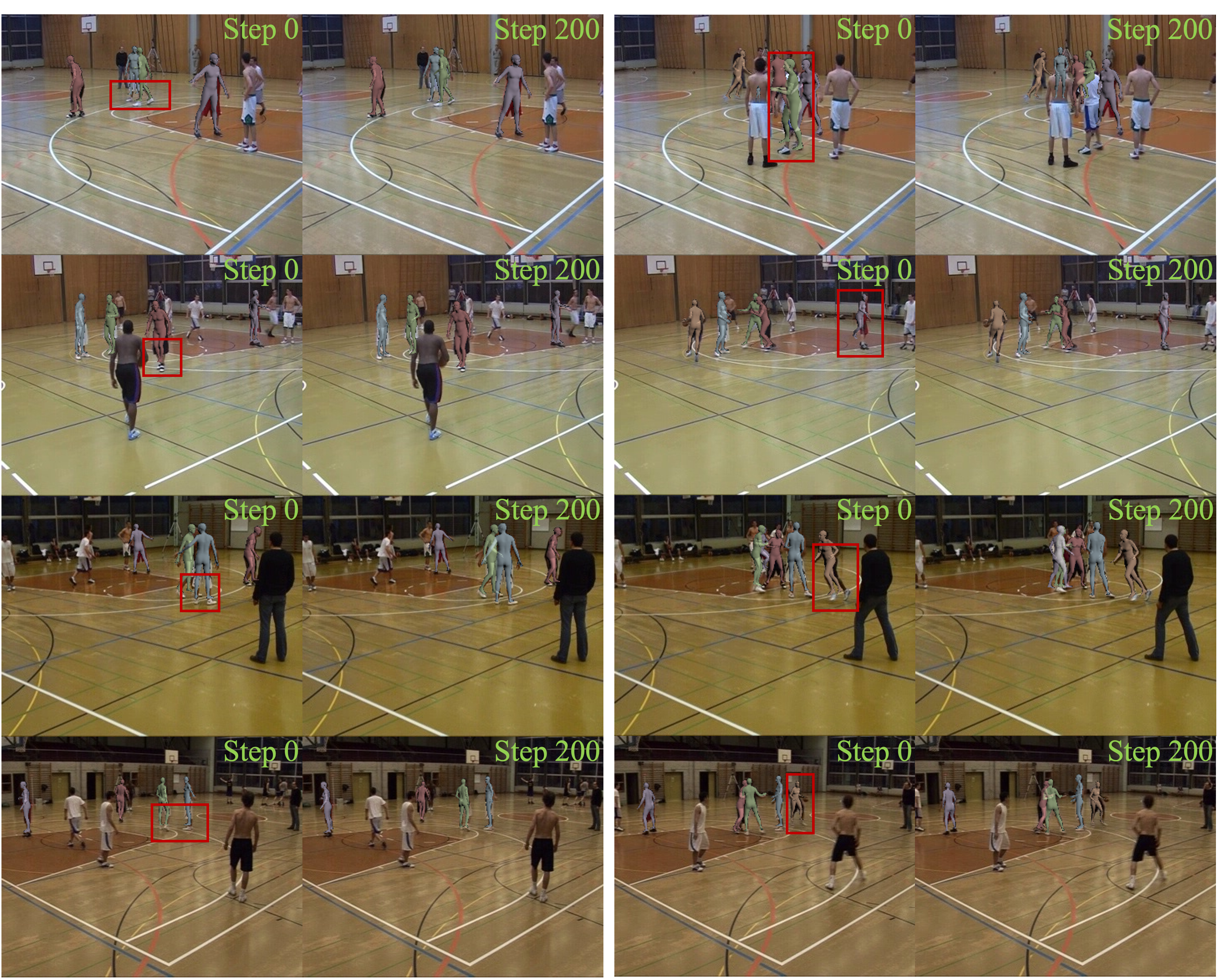}
    \caption{Mesh recovery and TTA alignment on real-world scenarios using the CVLab--EPFL Basketball Sequence. Red boxes highlight the misalignment in the initial predictions (Step 0).}
    \label{fig:sup_tta_basketball}
\end{figure*}

\cref{fig:sup_tta_basketball} shows qualitative results on a real-world scenario using the CVLab--EPFL~\cite{fleuret2008multicamera} Basketball Sequence. These results demonstrate the strong generalization capability of our approach beyond standard benchmark datasets. The red boxes highlight typical misalignment cases at Step 0 before TTA refinement. After optimization, the predicted meshes become significantly better aligned with the observed human poses, illustrating the robustness of our method in real-world multi-view settings.

\end{document}